\documentclass{article}

\usepackage[utf8]{inputenc} % allow utf-8 input
\usepackage[T1]{fontenc}    % use 8-bit T1 fonts
\usepackage{hyperref}       % hyperlinks
\usepackage{url}            % simple URL typesetting
\usepackage{booktabs}       % professional-quality tables
\usepackage{amsfonts}       % blackboard math symbols
\usepackage{nicefrac}       % compact symbols for 1/2, etc.
\usepackage{microtype}      % microtypography
\usepackage{graphicx}
\usepackage{subfigure}

\title{Facial Keypoints Detection}

\author{Shenghao Shi}
\date{}
\begin{document}

\maketitle

\begin{abstract}
  Detect facial keypoints is a critical element in face recognition. However, there is difficulty to catch keypoints on the face due to complex influences from original images, and there is no guidance to suitable algorithms. In this paper, we study different algorithms that can be applied to locate keyponits. Specifically: our framework (1)prepare the data for further investigation (2)Using PCA and LBP to process the data (3) Apply different algorithms to analysis data, including linear regression models, tree based model, neural network and convolutional neural network, etc. Finally we will give our conclusion and further research topic. A comprehensive set of experiments on dataset demonstrates the effectiveness of our framework.
\end{abstract}

\section{INTRODUCTION}
\subsection{Background}
Face recognition in unconstrained images is at the fore- front of the algorithmic perception revolution. The social and cultural implications of face recognition technologies are far reaching, yet the current performance gap in this domain between machines and the human visual system serves as a buffer from having to deal with these implications.

Recognizing faces is something that people usually do effortlessly and without much conscious thought, yet it has remained a difficult problem in the area of computer vision, where some 20 years of research is just beginning to yield useful technological solutions. As a biometric technology, automated face recognition has a number of desirable properties that are driving research into practical techniques.

The history of face recognition is as old as computer vision, both because of the practical importance of the topic and theoretical interest from cognitive scientists. Despite the fact that other methods of identification (such as fingerprints, or iris scans) can be more accurate, face recognition has always remains a major focus of research because of its non-invasive nature and because it is people’s primary method of person identification.

The problem of face recognition can be stated as ’identifying an individual from images of the face’ and encompasses a number of variations other than the most familiar application of mug shot identification. One notable aspect of face recognition is the broad interdisciplinary nature of the interest in it within computer recognition and pattern recognition; biometrics and security; multimedia processing; psychology and neuroscience. It is a field of research notable for the necessity and the richness of interaction between computer scientists and psychologists.

Perhaps the most famous early example of a face recognition system is due to Kohonen [3], who demonstrated that a simple neural net could perform face recognition for aligned and normalized face images. The type of network he employed computed a face description by approximating the eigenvectors of the face image’s autocorrelation matrix; these eigenvectors are now known as eigenfaces.

Kohonen’s system was not a practical success, however, because of the need for precise alignment and normalization. In following years many researchers tried face recog- nition schemes based on edges, inter-feature distances, and other neural net approaches. While several were successful on small databases of aligned images, none successfully addressed the more realistic problem of large databases where the location and scale of the face is unknown.

Kirby and Sirovich (1989) [2] later introduced an algebraic manipulation which made it easy to directly calculate the eigenfaces, and showed that fewer than 100 were required to accurately code carefully aligned and normalized face images. Turk and Pentland (1991) [1] then demonstrated that the residual error when coding using the eigenfaces could be used both to detect faces in cluttered natural imagery, and to determine the precise location and scale of faces in an image.

As the technology becomes mature, it has been widely in our daily life such as tracking faces in images and video analysing facial expressions detecting dysmorphic facial signs for medical diagnosis biometrics. I will list some of them in the next section.

In this article, we begin with detecting the location of keypoints on face images. This idea comes from our daily life because we can always identify our friends by their important feature on their face. It would be a good choice since this method really reduce the Computational complexity and could also achieve good Accuracy.

\subsection{Potential Applications}
Face recognition is closely related to many other domains, and shares a rich common literature with many of them. From our experiments, it would be clear that good testing performance could be achieved by predicting keypoint positions on face images. It provides us with a unique way to solve the problem.

For recognition of faces in video, face tracking is necessary, potentially in three dimensions with estimation of the head pose [15]. This naturally leads to estimation of the person’s focus of attention[16,17] and estimation of gaze [18] which are important in human computer interaction for understanding intention, particularly in conversational interfaces. This tack can be completed by predicting keypoint positions. Correspondingly there is much work on person tracking [19] and activity understanding [20] which are important guides for face tracking and for which face recognition is a valuable source of information.Recent studies have also begun to focus on facial expression analysis either to infer affective state [21] or for driving character animations particularly in MPEG-4 compression [22]. I believe that recent studies can be improved by this technology. The recognition of visual speech (i.e. lip-reading, particularly for the enhancement of acoustic speech recognition) is also a burgeoning face image processing area [23].

For example, The Face Recognition Technology (FERET) tests from Jonathan Phillips [26] provided an early benchmark of face recognition technologies.Phillips has continued the evaluation of face systems for US government agencies in the Face Recognition Vendor Tests [27]. This report provides an excellent independent evaluation of three state-of-the-art systems with concrete performance figures. The report highlights the limitations of current technology ,while under ideal conditions performance is excellent, under conditions of changing illumination, expression, resolution, distance or aging, performance falls off, in some cases dramatically. Current face recognition systems are not very robust yet against deviations from the ideal face image acquisition but there is continual performance improvement. For there published paper, the usage of getting keypoint positions on face images is obviously but not widely.

In addition, we can also apply this system to Access Control, Identification Systems, Surveillance, Pervasive Computing. Furthermore, the application should not be limited to face recognition. Although we will just discuss face recognition in this article, this idea of extracting several features to represent a complex picture may be widely used.
\section{REVIEW OF THE STATE-OF-THE-ART}
\subsection{hand-crafted features}
Most current face verification methods use hand-crafted
features. Moreover, these features are often combined to improve performance, even in the earliest LFW contributions. The systems that currently lead the performance charts employ tens of thousands of image descriptors [28,29,30]. In contrast, our method is applied directly to RGB pixel values, producing a very compact yet sparse descriptor.

\subsection{Deep neural nets}
Deep neural nets have also been applied in the past to face detection [31], face alignment [12] and face verification [32, 33]. In the unconstrained domain, Huang et al. [33] used as input LBP features and they showed improvement when combining with traditional methods. In our method we use raw images as our underlying representation, and to emphasize the contribution of our work, we avoid combining our features with engineered descriptors. We also provide a new architecture, that pushes further the limit of what is achievable with these networks by incorporating 3D alignment, customizing the architecture for aligned inputs, scaling the network by almost two order of magnitudes and demonstrating a simple knowledge transfer method once the network has been trained on a very large labeled dataset.

\subsection{Metric learning methods}
Metric learning methods are used heavily in face verification, often coupled with task-specific objectives [34, 10, 9]. Currently, the most successful system that uses a large data set of labeled faces [29] employs a clever transfer learning technique which adapts a Joint Bayesian model [9] learned on a dataset containing 99,773 images from 2,995 different subjects, to the LFW image domain. Here, in order to demonstrate the effectiveness of the features, we keep the distance learning step trivial.

\subsection{Face Detection}
Naturally, before recognizing a face, it must be located in the image. This topic is very important and popular among researchers in recent years. In some cooperative systems, face detection is obviated by constraining the user. Most systems use a combination of skin-tone and face texture to determine the location of a face and use an image pyramid to allow faces of varying sizes to be detected. Increasingly, systems are being developed to detect faces that are not fullfrontal [36]. Cues such as movement and person detection can be used [35] to localize faces for recognition. Typically translation, scale and in-plane rotation for the face are estimated simultaneously, along with rotation-in-depth when this is considered.

\subsection{Face Recognition}
There is a great diversity in the way facial appearance is interpreted for recognition by an automatic system. Currently a number of different systems are under development, and which is most appropriate may depend on the application domain. A major difference in approaches is whether to represent the appearance of the face, or the geometry. Brunelli and Poggio [37] have compared these two approaches, but ultimately most systems today use a combination of both appearance and geometry. Geometry is difficult to measure with any accuracy, particularly from a single still image, but provides more robustness against disguises and aging. Appearance information is readily obtained from a face image, but is more subject to superficial variation, particularly from pose and expression changes. In practice for most purposes, even appearance-based systems must estimate some geometrical parameters in order to derive a ’shape- free’ representation that is independent of expression and pose artefacts [38, 39]. This is achieved by finding facial landmarks and warping the face to a canonical neutral pose and expression. Facial features are also important for geometric approaches and for anchoring local representations.

\section{OBJECTION AND CONTRIBUTIONS}
The goal of the project is to locate specific keypoints on face images.And we will build an algorithm, that, given an image of a face, automatically locates where these keypoints are located.

In our project, we try exploiting machine learning algorithms taught in the course, including linear regression, KNN, logistic regression, neutral network as well as tree- based methods. Of course, all above algorithms we apply in practice are quite basic, and have been developed relatively mature especially the Convolutional Neural Network, namely CNN. However, our contributions are that we compare these methods and explain the advantage and disadvantage of using these methods for the project problem. What’s more, we spend a lot of time on data preprocessing. Because we think in the case where the methods have been fixed, the better the data is process, the better result we can get. This work is unique and exactly important. And these ideas are likely to be used by other people to better their experimental results. What’s more, in the end of our article, we put forward some further improvement which may inspires new ideas with regard to the detection of keypoints.

In this experiment, the main language we use is $Python$.

In addition, we use these packages to achieve our algorithm:

(1)Numpy

(2)Scipy

(3)Scikit

(4)Theano

(5)Keras

Let’s start by describing some of the work that we have done.

\section{DATA PREPROCESSING}

\subsection{original data}
Each predicted keypoint is specified by an (x,y) real-valued pair in the space of pixel indices. There are 15 keypoints, which represent the following elements of the face:

$left_eye_center, right_eye_center,$

$left_eye_inner_corner, left_eye_outer_corner, $

$right_eye_inner_corner, right_eye_outer_corner, $

$left_eyebrow_inner_end, left_eyebrow_outer_end, $

$right_eyebrow_inner_end, right_eyebrow_outer_end, $

$nose_tip,$

$mouth_left_corner, mouth_right_corner, $

$mouth_center_top_lip, mouth_center_bottom_lip$

Left and right here refers to the point of view of the subject.

In some examples, some of the target keypoint positions are missing (encoded as missing entries in the csv, i.e., with nothing between two commas).

The input image is given in the last field of the data files, and consists of a list of pixels (ordered by row), as integers in (0,255). The images are 96x96 pixels.

Then we download the data files from Kaggle websites:

training.csv: list of training 7049 images. Each row contains the (x,y) coordinates for 15 keypoints, and image data as row-ordered list of pixels.

test.csv: list of 1783 test images. Each row contains ImageId and image data as row-ordered list of pixels.

\subsection{spilt the data}
We spilt the $training.csv$ into $keypoint.csv$ and $im.csv$. Each row of $keypoint.csv$ contains the (x,y) coordinates for 15 keypoints, while that of $im.csv$ contains the image data as row-ordered list of pixels.

To check the data integrity, we have a close-up view of the data stored in $keypoint.csv$, and find the dataset has some problems. It is two different datasets put together into one for this competition, i.e., the target keypoints positions of the first 2284 examples are almost complete and the latter 4765 examples only contain the (x,y) coordinates for 4 keypoints while the other 11 keypoints positions are all missing. To illustrate, a sketch map is displayed below:

\begin{figure}
\centering
\includegraphics[scale=0.6]{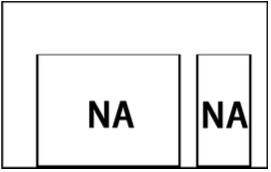}
\caption{Sketch Map}
\end{figure}

So it is important to split the data and train seperate models. We firstly split it into one dataset with 15 keypoints, i.e. section A, and one with 4 keypoints, i.e. section B, just as illustrated below:

\begin{figure}
\centering
\includegraphics[scale=0.6]{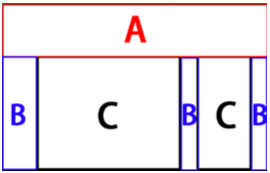}
\caption{One Way to Spilt Data Set}
\end{figure}

However, at the half-way through our project, we come up with a better way to spilt the data. Actually, after visualizing the image data in $im.csv$, we find there is no significant difference between the data corresponding to section A and section B. Therefore, we just spilt the data into one dataset with 4 keypoints, i.e. section D, and one with 11 keypoints, i.e. section E, which is illustrated as follow:

\begin{figure}
\centering
\includegraphics[scale=0.6]{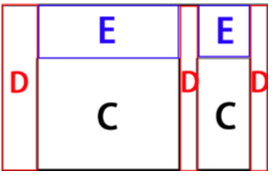}
\caption{One Way to Spilt Data Set}
\end{figure}

As for the new way to spilt the data, there are 7049 examples in $keypoint_4f.csv$ and $im_4f.csv$, 2284 examples in $keypoint_11f.csv$ and $im_11f.csv$. Compared to the former way, we just train a model for the 4 keypoints, and the other for the rest 11 keypoints.

In this way, we can make a better prediction for these 4 keypoints. Of course, we need to notice there is some inter- relation among the 15 keypoints position, for example, the nose is likely to fall on the perpendicular bisector of the line-segment connecting two eyes with good probability. Considering the interrelation, the prediction accuracy for the rest 11 keypionts may slightly decline for the absence of the 4 keypoints. But in our models, i.e. linear regression, K- NN, logistic regression, SVM, Neural network, we seldom make use of the inter-relation among the 15 keypoints, thus, the second way is more reasonable. Keep in mind when the inter-relation is included in the model, the first way to spilt the data may be better.

Take a notice that, apart from section C, there are also few target keypoint points are missing in section A and B. In order to facilitate subsequent analysis process, we substitude the missing data with the corresponding column mean.

Moreover, since we don’t have the ground-truth for $test.csv$, for the paper writing purposes, we split the training set into 90\% and 10\% respectively. And the 10\% held-out data can be served as the testing data to evaluate the algorithms. In other words, $test.csv$ is put aside in our project. Therefore, we divide the

$keypoint_4f.csv$ into $keypoint_train_4f.csv$ and $key-point_test_4f.csv$;

$im_4f.csv$ into $im_train_4f.csv$ and $im_test_4f.csv$;

$keypoint_11f.csv$ into $keypoint_train_11f.csv$ and $key- point_test_11f.csv$;

$im_11f.csv$ into $im_train_11f.csv$ and $im_test_11f.csv$.
\subsection{Visualize the data}
For better understanding, we visualize the first image in $im_train_4f.csv$. We first reshape these 9216 integers into a 96x96 matrix, then use R's image function:

\begin{figure}
\centering
\includegraphics[scale=0.6]{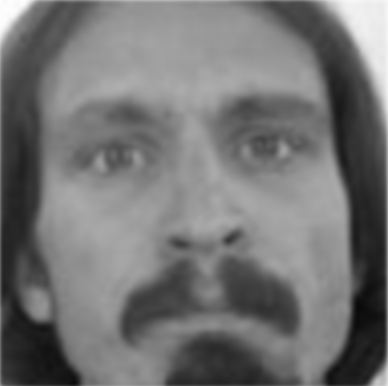}
\caption{the first image}
\end{figure}

We can the add some keypoints(from the corresponding file $keypoint_train_4f.csv$) to check if everything is correct so far. Let’s color the coordinates for the eyes and nose as an example:

\begin{figure}
\centering
\includegraphics[scale=0.6]{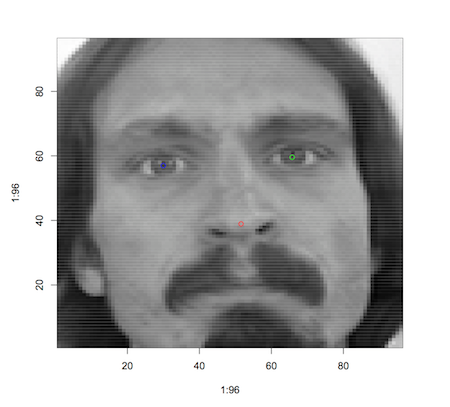}
\caption{add keypoints to the image}
\end{figure}

We can see the positions of the centers of the nose and two eyes are all labelled correctly.

Another good check is to see how variable is our data. For example, where are the nose centers in the 7049 images?

\begin{figure}
\centering
\includegraphics[scale=0.6]{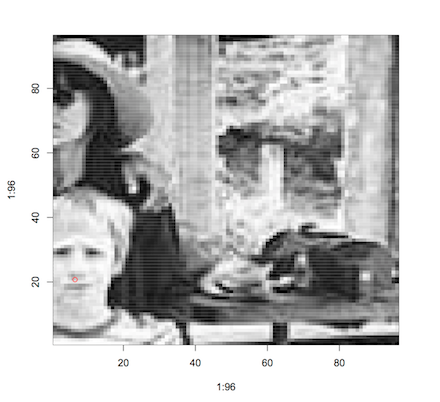}
\caption{an extreme example}
\end{figure}

We can see from above, most nose points are concentrated in the central region(as expected), but there are quite a few outliers that deserve further investigation, as they could be labeling errors. Looking at one extreme example:

\begin{figure}
\centering
\includegraphics[scale=0.6]{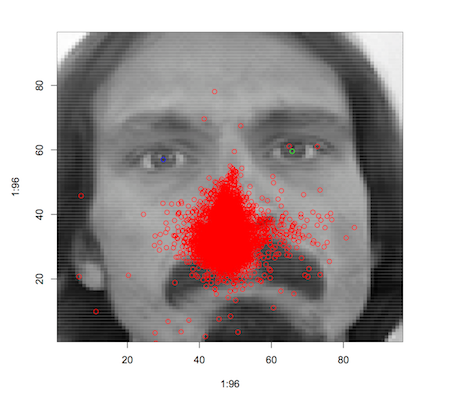}
\caption{centers of each nose in the 7049 images}
\end{figure}

In this case there’s no labeling error, but this shows that not all faces are centralized as one might expect.

\subsection{Face Detector}
Just as Fig.7 indicates, not all faces are centralized, thus we decide to build a face detector first to find bounding boxes for faces in the image. We get the result as follow via opencv:

\begin{figure}
\centering
\includegraphics[scale=0.6]{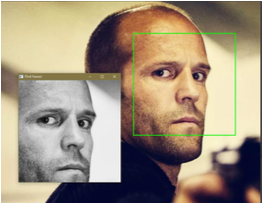}
\caption{find bounding box for face}
\end{figure}

We can see from above the method works quite well. However, when we apply the method to the training data, the result is really upset:

\begin{figure}
\centering
\includegraphics[scale=0.6]{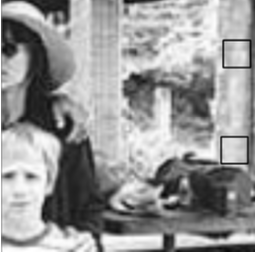}
\caption{find bounding box for face}
\end{figure}

This is because the size of images data is 96x96, which indicates the images are badly out of focus.

To get a better result, we may train a neural network separately to detect the face in the image, which need further exploration. Or otherwise, we can use opencv to detect the keypoints of the image directly. However, opencv doesn’t utilize the training dataset and it’s a mature model that has been trained. Therefore, this method is of little significance as to our project.

\subsection{LBP}
LBP (Local Binary Pattern) is an operator used to describe the local texture features of images. It has the advantages of rotation invariance and gray invariance. It was first proposed by T. Ojala, M. Pietikainen, and D. Harwood in 1994 for texture feature extraction. Moreover, the extracted features are local texture features of the image.It has since been found to be a powerful feature for texture classification; it has further been determined that when LBP is combined with the Histogram of oriented gradients (HOG) descriptor, it improves the detection performance considerably on some datasets. A comparison of several improvements of the original LBP in the field of background subtraction was made in 2015 by Silva et al. A full survey of the different versions of LBP can be found in Bouwmans et al..

\subsubsection{Description of LBP features}
The original LBP operator is defined as a 3 * 3 window, and the window center pixel is set as a threshold, compared to its 8 neighbors (on its left-top, left-middle, left-bottom, right-top, etc.). If the neighbor’s value is greater than the center pixel value, then write "1", otherwise write "0". In this way, we can obtain an 8-digit binary number (which is usually converted to decimal for convenience), which is the LBP value of the center pixel. This value is always used to reflect the texture information for this area.
For better understanding, we give a simple case to calculate LBP value just as shown below:

\begin{figure}
\centering
\includegraphics[scale=0.6]{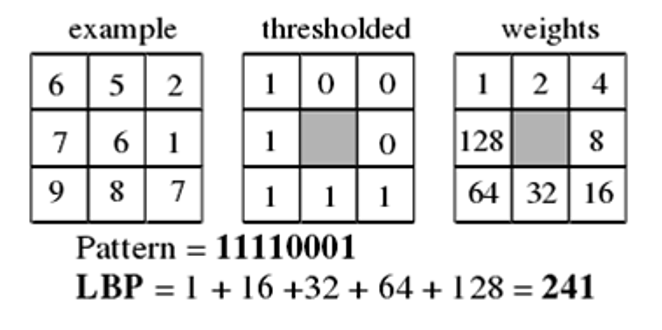}
\caption{Calculation of LBP}
\end{figure}

So that the pixel value of center point is replaced by 241. Note that it doesn’t mandate a particular order in which LBP value is calculated. We only need to maintain the same order in the same process.

The calculation formula of LBP is
\begin{displaymath}
LBP=\sum_{p=0}^{P-1}s(g_{p}-g_{c})2^{p}
\end{displaymath}

\begin{displaymath}
s(x)=\left \{ \begin{array}{ll} 
1 & x \ge 0 \\
0 & x<0 
\end{array} \right.
\end{displaymath}

where $g_{p}$ is the neighbor's pixel value and $g_{c}$ is the center pixel value 

After the original LBP was put forward, the researchers constantly propose a variety of improvements and optimization.

\subsubsection{Improved version of LBP}
(1)Circular LBP Operator

The main drawback of the basic LBP operator is that it only covers a small area with a fixed radius, which obviously can’t meet the needs of different size and frequency texture. In order to adapt to different scale of texture features as well as achieve the requirements of gray and rotation invariance, Ojala et al. improves the LBP operator: extend the 3x3 neighborhood to any size of neighborhood, and replace the square neighborhood with a circular neighborhood. The improved LBP operator allows any number of pixels in a circular neighborhood with the radius R.

\begin{figure}
\centering
\includegraphics[scale=0.6]{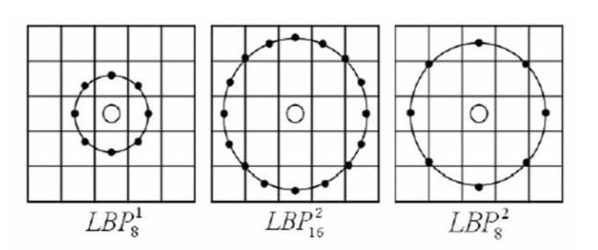}
\caption{Circular LBP Operator}
\end{figure}

Note: $LBP_{p}^{r}$, p is the number of sampling pixels; r is the radius of circular neighborhood.

(2)LBP Rotation Invariant Mode

As can be seen from the definition of LBP, LBP operator is gray-invariant, but not rotation-invariant. Rotation of the image can result in different LBP values.

Maenpaa et al. extends the LBP operator and proposed a rotation invariant LBP operator, that is, rotating the circular neighborhood to obtain a series of initially defined LBP values, then taking the minimum value as the LBP value of the neighborhood.

Figure 3 shows the diagram of obtaining the rotation in- variant LBP. The number below the operator represents the corresponding LBP value. After the treatment of rotation, the final rotation-invariant LBP value is 15. In other words, the 8 LBP modes shown in the figure correspond to the same rotation-invariant LBP pattern 00001111.

\begin{figure}
\centering
\includegraphics[scale=0.6]{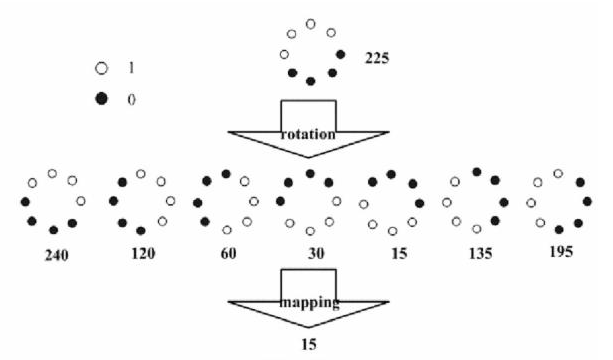}
\caption{Diagram of Rotation Invariant LBP}
\end{figure}

Of course, there are still many other improvements of LBP, but we’ll not described in this report. Interested people can learn more about it by yourselves.

\subsubsection{Steps of Extracting the LBP Feature Vector}

The LBP feature vector, in its simplest form, is created in the following manner:
\begin{itemize}
\item Divide the examined window into cells (e.g. 16x16 pixels for each cell).

\item For each pixel in a cell, compare the pixel to each of its 8 neighbors (on its left-top, left-middle, left- bottom, right-top, etc.). Follow the pixels along a circle, i.e. clockwise or counter-clockwise.

\item Where the center pixel’s value is greater than the neighbor’s value, write "0". Otherwise, write "1". This gives an 8-digit binary number (which is usually converted to decimal for convenience).

\item Compute the histogram, over the cell, of the fre- quency of each "number" occurring (i.e., each com- bination of which pixels are smaller and which are greater than the center). This histogram can be seen as a 256-dimensional feature vector.

\item Optionally normalize the histogram.

\item Concatenate (normalized) histograms of all cells. This gives a feature vector for the entire window.
\end{itemize}
The feature vector can now be processed using the SVM or some other machine-learning algorithm to classify images. Such classifiers can be used for face recognition or texture analysis.

It is obvious that the LBP operator can get a LBP value at each pixel point. Then, after extracting the LBP feature vector for an image(each pixel records the gray value), the LBP feature we get is still a "picture" (each pixel records the LBP value) just as below:

\begin{figure}
\centering
\includegraphics[scale=0.6]{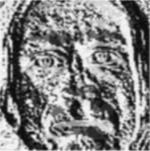}
\caption{post\_LBP}
\end{figure}

\subsection{PCA}

In our project, the dimension of the image data is 9216 (96x96), which is rather large and adds to the difficulty of data analysis. And we naturally think of dimensionality reduction because of its advantages in the case of processing high-dimensional data:

\begin{itemize}

\item low-dimensional data is easier to handle;

\item correlation features of the data set, especially the important features can stand out;

\item remove the data noise;

\item reduce the cost of the algorithm and more conducive to solving the problem when computational resources are limited;

\end{itemize}

PCA, Factor Analysis and Independent Component Analysis are the common algorithms for dimensionality reduction. Among them, PCA is the most widely used method and is also adopted by us.

\subsubsection{Principle of PCA}

PCA is a statistical procedure that uses an orthogonal transformation to convert a set of observations of possibly correlated variables into a set of values of linearly uncorre- lated variables called principal components. The number of principal components is less than or equal to the number of original variables. This transformation is defined in such a way that the first principal component has the largest possi- ble variance (that is, accounts for as much of the variability in the data as possible), and each succeeding component in turn has the highest variance possible under the constraint that it is orthogonal to the preceding components. The resulting vectors are an uncorrelated orthogonal basis set. And PCA is sensitive to the relative scaling of the original variables.

\begin{figure}
\centering
\includegraphics[scale=0.6]{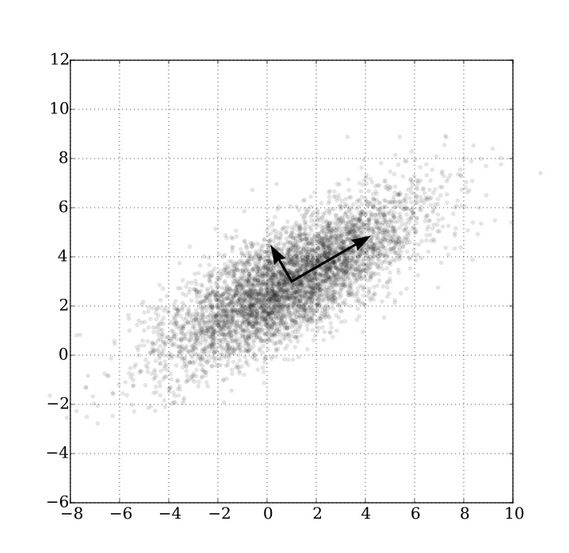}
\caption{Principle Sketch of PCA}
\end{figure}

\subsubsection{Implementation Process of PCA}

On the basis of linear algebra, we can obtain the values of these principal components by analysing the covariance ma- trix of the data set and its eigenvalue. Once the eigenvectors are obtained, we can retain the largest N vectors. The data set can then be converted to the new space by multiplying these N eigenvectors.

More detailly, we can describe the implementation pro- cess of PCA as follows:

\begin{itemize}

\item normalize the data;

\item obtain the covariance matrix;

\item calculate the eigenvalues and eigenvectors of the covariance matrix;

\item sort the eigenvalues from large to small;

\item retain the largest N eigenvectors;

\item convert the data into the new space constructed by the above N eigenvectors.

\end{itemize}

It should be stressed that the retained eigenvectors need to cover more than 90\% of the information. And in practical application, actually, we retain the largest 256 eigenvectors, which cover more than 95\% of the information.

\section{ALGORITHMS AND ANALYSIS}

In this part, we will talk about the algorithms we used to recognize the keypoints on the faces and their task completions in order to explore their differences and find which would be the most suitable one. In addition, all the codes are also provided in case you want to run it by yourselves. However, please note that there is randomness among different experiments, which means you should not be surprised if you cannot obtain the same result.

In this section, I just focus on the specific algorithm itself. And in the next section, the contrast among different algorithms will be proposed. And we will describe our step in P ython to build Neural network and Convolutional neural network since they are the most challenging aspect in the experiment.

First of all, I list all the results here in tables:

\begin{tabular}{|c|c|c|}
\hline
&RMSE1&RMSE2 \\
\hline
Knn&3.375&2.346 \\
\hline
Linear & 4.513&6.020 \\
\hline
Lasso &3.558 & 2.979 \\
\hline
Elastic & 4.044 &2.959 \\
\hline
Ridge & 8.464 &2.609 \\
\hline
Decision tree &3.745 &4.101 \\
\hline 
Neural network & 2.923 &2.875 \\
\hline 
CNN & 1.972 &2.086 \\
\hline
\end{tabular}

\subsection{Knn}

We prefer to start our discussion among the algorithms from kNN(k-NearestNeighbor) because it is very simple to understand and visualize.

KNN regression can be used in cases where the data la- bels are continuous rather than discrete variables. The label assigned to a query point is computed based the mean of the labels of its nearest neighbors. The advantages lie in its simplicity and effectiveness. Theoretically, the more training instances we can provide, the better the performance of Knn algorithms will be. Luckily, there is a big data set for us to train so RMSE1 has been reduced to about 3.375 with $k=5$.

Talking about the weak points about Knn. I want to mention the Curse of Dimensionality, which significantly cut down the power of kNN algorithm. In this case, this problem is obvious because we have 30 and 8 dimensions in each data sets. Although the huge number of instances reduce the Curse of Dimensionality, to some extent, it is still rigorous.

However, Knn receives 2.346 RMSE2 on a the data set of 8 dimensions while 3.375 RMSE1 on a data set of 30 dimensions . It uncovers the Curse of Dimensionality, to some extent.

\subsection{linear regression}

We fit a linear model with coefficients $w=(w_{1}, \ldots , w_{p})$ to minimize the residual sum of squares between the observed responses in the dataset, and the responses predicted by the linear approximation. Mathematically it solves a problem of the form
\begin{displaymath}
min || \mathbf{X} w - y||_{2}^{2}
\end{displaymath}

where $\mathbf{X}$ stands for the observed variables in the dataset and $y$ represents the responses.

Obviously, linear regression is too simple to solve the problems in real world unless the relationships between observed variables and responses are linear itself. However, there is no evidence to prove this kind of bind. Therefore, we received $8.4634$ RMSE1, which is the worst among all the methods we used in the experiment.

Considering another reason of high RMSE, we think the problem of overfitting is the biggest problem. In the following section, you will see how Lasso regression and Elastic Net regression improve the RMSE by overcoming the problem of too much parameters.

\subsection{Lasso}
The Lasso regression is a linear model that estimates sparse coefficients. It is useful in some contexts due to its tendency to prefer solutions with fewer parameter values, effectively reducing the number of variables upon which the given solution is dependent. For this reason, the Lasso and its variants are fundamental to the field of compressed sensing. Under certain conditions, it can recover the exact set of non-zero weights.

Mathematically, it consists of a linear model trained with $\ell_{1}$ prior as regularizer. The objective function to minimize is:
\begin{displaymath}
min \frac{1}{2n_{samples}} ||\mathbf{X} w- y||^{2}_{2} +\alpha ||w||_{1}
\end{displaymath}

And it can be proved that solving for the lasso regression is equivalent to solve the following problem:
\begin{displaymath}
min \sum_{i=1}^{n} (y_{i} -\beta_{0} - \sum_{j=1}^{p} \beta_{j} x_{ij})^{2} ,  \sum_{j=1}^{p} |\beta_{j}| \le s
\end{displaymath}

Lasso regression enjoys a good RMSE1 with only 3.559. Some parameters may be extra since sometimes we do not need such information to recognize a face. On the other hand, parameter redundancy may cause the problem of overfitting. Lasso regression has a the ability to overcome overfitting and that is the reason why it is so popular among machine learning skills.

The choice of the $\alpha$ is very important because large $\alpha$ will squeeze almost all the parameters to zero, which obviously cannot meet our requirements, while small $\alpha$ does not have the ability to fulfill the task as a lasso regression, in contrast, it may perform as a normal linear regression.

The best way to choose $\alpha$ is cross validation. However, cross validation is not easily to do on this dataset because it is too much big so that long times would be taken to run the code. Luckily, we almost find the best choice ( $\alpha$ =0.1). However, you should note that since there is no Theoretical guidance to $\alpha$, so this may not be the unique choice.

\subsection{Elastic Net}

Elastic Net is a linear regression model trained with $\ell_{1}$ and $\ell_{2}$ prior as regularizer. This combination allows for learning a sparse model where few of the weights are non- zero like Lasso, while still maintaining the regularization properties of Ridge.

Elastic-net is useful when there are multiple features which are correlated with one another. Lasso is likely to pick one of these at random, while elastic-net is likely to pick both.

A practical advantage of trading-off between Lasso and Ridge is it allows Elastic-Net to inherit some of Ridge’s stability under rotation.

The objective function to minimize is in this case:
\begin{displaymath}
min \frac{1}{2n_{samples}} || \mathbf{X} w-y||^{2}_{2} +\alpha \rho ||w||_{1} + \frac{\alpha(1-\rho)}{2} ||w||_{2}^{2}
\end{displaymath}

In elastic net there is two parameters to be adjusted, which significantly increases the difficulty of tuning. In our experiment the RMSE1 is 4.04. Therefore, we prefer to attribute the poor performance to parameters. Although we determine the parameters using cross validation, it is still hard to tune. Finally, the parameters are fixed with
$\alpha=0.1, \rho=0.5$.

Compared with Lasso regression, Elastic Net achieves a higher test error. However, we cannot conclude that Elastic Net cannot compare with the Lasso regression. Instead, Lasso regression is strong enough to prevent the system from overfitting rather than.

\subsection{Ridge Regression}

Compared with linear regression ridge regression balance the error and variance by adding penalty. The parameters of ridge regression are obtained my minimize:
\begin{displaymath}
\sum_{i=1}^{n}(y_{i}-\beta_{0}-\sum_{j=1}^{p} \beta_{j} x_{ij})^{2} _\lambda \sum_{j=1}^{p} \beta_{j}^{2}
\end{displaymath}

If $\lambda=0$ it will degenerate into normal linear regression. In addition, it can be proved that solving for ridge regression is equivalent to solving the following problem:
\begin{displaymath}
min \sum_{i=1}^{n} (y_{i} -\beta_{0} -\sum_{j=1}^{p} \beta_{j}x_{ij})^{2}, \sum_{j=1}^{p} \beta_{j}^{2} \le s
\end{displaymath}

Ridge regression has closed form solution as a convex optimization problem. This property improved the speed of the code largely. However, ridge regression has huge gap between RMSE1 and RMSE2, as can be seen from the table above. This problem confused us a lot during the experiment. However, after dividing the data set again, we obtain a normal result for ridge regression with RMSE1 = 2.967 and RMSE2 = 3.104. In the table above we still report the previous RMSE because we want to mention this surprising result may due to luck but still need further investigation.

\subsection{Decision tree}

Decision Trees (DTs) are a non-parametric supervised learning method used for classification and regression. The goal is to create a model that predicts the value of a target variable by learning simple decision rules inferred from the data features.

Some advantages of decision trees are:

\begin{itemize}

\item Simple to understand and to interpret. Trees can be visualized.

\item Requires little data preparation. Other techniques often require data normalization, dummy variables need to be created and blank values to be removed. Note however that this module does not support missing values.

\item Use a white box model. If a given situation is ob- servable in a model, the explanation for the condition is easily explained by boolean logic. By contrast, in a black box model results may be more difficult to interpret.

\end{itemize}

The disadvantages of decision trees include:

\begin{itemize}

\item Decision-tree learners can create over-complex trees that do not generalize the data well. This is called over- fitting. Mechanisms such as pruning (not currently sup- ported), setting the minimum number of samples required at a leaf node or setting the maximum depth of the tree are necessary to avoid this problem.

\item Decision trees can be unstable because small variations in the data might result in a completely different tree being generated. This problem is mitigated by using decision trees within an ensemble.

\end{itemize}

The RMSE1 of decision trees is 3.75 with $k = 5$. We have the confidence to believe that overfitting would not occur with applying only $k = 5$ to such a big data set. However, decision tree may ignore the correlation between characters, which is an important property in this experiment.

\subsection{Neural network}

\subsubsection{Package description}

As you know Keras is a high-level neural networks library, written in Python and capable of running on top of either TensorFlow or Theano. It was developed with a focus on enabling fast experimentation. Being able to go from idea to result with the least possible delay is key to doing good research.

In this experiment, we first study how to use the pack- ages Keras and Theano to build our neural network, then we choose two different neural networks to achieve our goal.They have a common feature is very precise, but the calculation is very large, we spent 5 days to debug them. Unfortunately, BP neural network has not reached an ideal result, but CNN’s result is very good.

\subsubsection{Steps to build neural network}

Now, we’ll give you an idea of how to build the BP neural network.

This has the same idea as handwritten number recognition, but is slightly different in some detail. This is a regression problem, not a classification problem. First we’d better to use the ’MSE’ to replace the ’categorical crossentropy’. Second, we must use the linear function $f(x)=x$ as the activation function in the last layer, because the value of the other functions will be limited to 0 and 1, can not complete the predict. The other things we have done as follow:

1 We choose the different optimizer to build our neural network, such as ’Rmsprop’ and ’SGD’.

2 We use 4 layers which has 300,150,50,8 neural for training.

3 After some times attempt, we choose the nb\_epoch=500, batch\_size=30.

4 we also use the dropout idea to get a model with stronger generalization ability.

\subsubsection{Analysis}

First, we construct $M$ linear combinations of the input variables $x_{1},\ldots, x_{D}$ in the form 
\begin{displaymath}
a_{j} = \sum_{i=1}^{D} w_{ji}^{(1)} x_{i}+ w_{j0}^{(1)}
\end{displaymath}

where $j=1,\ldots, M$ and the superscript (1) indicates that the corresponding parameters are in the first ’layer’ of the network. The quantities $a_{j}$ are known as activations. Each of them is then transformed using a differentiable, nonlinear activation function $h(x)$ to give

\begin{displaymath}
z_{j}=h(a_{j})
\end{displaymath}

The nonlinear function $h(x)$ are generally chosen to be sigmoidal functions such as the logistic sigmoid or the ’tanh’ function.

Following these steps, we can extend the neural network to any hidden units and any layers we want.

The method we used to train the network is Backpropagation, I won’t list the details mathematically, you can refer to [40] and check our code.

Neural network has the advantage of high classification rate and good learning skills. All the knowledges are evenly distributed in the system so that neural that can approximate any nonlinear system and it also has the ability of fault tolerance.

However, a lager number of parameters are needed for neural network. And there is no theoretical guidance to the choice of layers and hidden units, including other important parameters, which means that tuning is a pivotal step to a perfect neural network.

In the experiment we only receive 2.923 RMSE1, which is a terrible result obtained from neural network. Tuning must be one of the reasons but a further research should be requested.

\subsection{Convolutional Network Network}

\subsubsection{Steps to build convolutional neural network}

As we all know, the most popular technique in image recognition is CNN. So we spend the longest time in this method, now I will introduce the idea of our CNN.

\begin{itemize}

\item We use PCA method to reduce dimension from $96 \times 96$ to $16 \times 16$, This method saves 95\% information, and makes the calculation faster.

\item Because Keras requires the input of 4-dimensional vectors, so we need replace the $7049 \times 16 \times 16$ image as $7649 \times 16 \times 16 \times 1$ image.

\item We use 2 convolution layers. The first layer we use 32 filters with dimension $5 \times 5$, then the pool\_size is $2 \times 2$. In the second layer, we use the 8 filter with dimension $3 \times 3$, then the pool\_size is $2 \times 2$. we also use the dropout method in each layer.

\item After the convolution layer, we flatten it and add a normal hidden layer with 100 neural.

\item Optimizer and loss function is similar with BP network.

\item After some times attempt, we choose the nb\_epoch=400, batch\_size=50.

\end{itemize}

We try to install the cuda and use the GPU to compute it but fail, so we only use the CPU to train this neural network,it takes a very long time. Each test time are more than 10 hours, so the number of adjusting the parameters is not enough. But fortunately, the results are pretty good!

Following is a picture of a small demo run on our own computer. However, the whole code should be run on the server.

\begin{figure}
\centering
\includegraphics[scale=0.6]{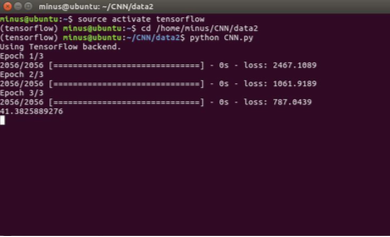}
\caption{Principle Sketch of PCA}
\end{figure}

\subsubsection{Analysis}

The most successful approach is the convolutional neural network, which has been widely applied to image data. Convolutional neural network has three important mechanisms: (i) local receptive fields, (ii) weight sharing, and (iii)subsampling. The structure of a convolutional network is illustrated below:

\begin{figure}
\centering
\includegraphics[scale=0.6]{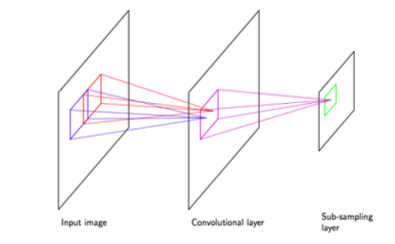}
\caption{Principle Sketch of PCA}
\end{figure}

In the convolutional layer the units are organized into planes, each of which is called a feature map. Units in a feature map each take inputs only from a small subregion of the image, and all of the units in a feature map are considered to share the same weight values. For instance, a feature map might consists of 100 units arranged in a $10 \times 10$ grid, with each unit taking inputs from a $5 \times 5$ pixel patch of the image. The whole feature map therefore has 25 adjustable weight parameters plus one adjustable bias parameter. Input values from a patch are linearly combined using the weights and the bias, and the result transformed by a sigmoidal nonlinearity.

The whole network can be trained by error minimization using backpropagation to evaluate the gradient of the error function. This involves a slight modification of the usual backpropagation algorithm to ensure that the shared-weight constraints are satisfied. Due to the use of local receptive fields, the number of weights in the network is smaller than if the network were fully connected. Furthermore, the number of independent parameters to be learned from the data is much smaller still, due to the substantial numbers of constraints on the weights.

Convolutional neural network is an excellent approach to image recognition because it makes the full use of local feature and shares soft weights, which is critical to images.

CNN achieves the best result with 1.972 RMSE1 and 2.086 RMSE2.

\section{RESULTS OPTIMIZING}
We re-run our codes on the pre-processed data by means of LBP and PCA, and get the better result:

\begin{tabular}{|c|c|c|}
\hline
&RMSE1&RMSE2 \\
\hline
Knn&2.900&2.256 \\
\hline
Linear & 3.011&2.456 \\
\hline
Lasso &3.009 & 2.352 \\
\hline
Elastic & 3.010 &2.354 \\
\hline
Ridge & 3.011 &2.356 \\
\hline
Decision tree &3.827 &2.354 \\
\hline 
Neural network & 2.498 &2.432 \\
\hline 
CNN & 1.909 &1.874 \\
\hline
\end{tabular}

\subsubsection{Analysis of PCA}

The transformation of PCA is to project the original sample data (n-dimensions) into a k-dimensional(k<n) orthogonal coordinate system, discarding the information in other dimensions. In this way, PCA can extract the main influencing factors, revealing the properties of the data.

What’s more, in fact, the variables in image data have a quite strong internal correlation. For instance, eyes symmetrically distribute on both sides of the nose. However, in our models, we ignore this correlation subconsciously, which makes the RMSE not very good.

But when we adopt PCA, we can remove the correlation among different variables because the project space is an orthogonal coordinate system. And the running results also strongly support this conclusion.

By the way, for PCA is a dimension-reduction algorithm, it will discard part of the information inevitably, but it does’t hurt the important essentials. Because we retain 95\% of the information in practice.

We originally didn’t expect PCA to behave well, instead, we just want to reduce the time for running our code. Surprisingly, PCA reduces all the RMSE without exception. After the gun, we may conclude that ’Curse of Dimensionality’ is a serious problem in this experiment. PCA has the ability to overcome overfitting and balance bias-variance.

\subsubsection{Analysis of LBP}

Facial features vary greatly from one individual to another, and even for a single individual, there is a large amount of variation due to 3D pose, size, position, viewing angle, and illumination conditions. LBP allows you to capture the details of the image very well. In fact, researchers can achieve the state-of-the-art level in texture classification by using it. Actually, LBP is kind of prudent with the monotonous change in gray-scale. We can see the LBP images of the same photo under different illumination conditions:

\begin{figure}
\centering
\includegraphics[scale=0.4]{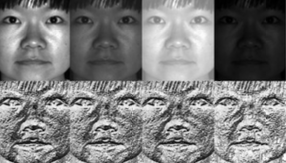}
\caption{images under different illumination conditions}
\end{figure}

We can judge that LBP feature is not sensitive to light, in other words, it can eliminate the impact of illumination conditions effectively when we are locate the position of keypoints. Thus, there is no surprise that LBP data can reduce the root mean squared error significantly.

\section{COMPARISON OF VARIOUS METHODS AND FURTHER IMPROVEMENT}

In this section, we will focus on the differences among all the proposed methods above.

First of all, we will begin with convolutional neural network.(CNN) There is no doubt that CNN is the best solution to this problem with such a low RMSE. It has been proved that CNN performs well to image recognition both from theory and practice. The advantages include feature extraction and soft weight sharing have been discussed before, which is the unique properties of CNN. It is these characters that make CNN such a perfect approach. However, the disadvantage still need to be mentioned since this cannot be felt from these paper unless to run our code on computers. Very long time will be taken though the code is run on the server. It raises the problem of tuning because it is time consuming. Thence, we have reasons to believe that better RMSE can be achieved if suitable parameters are applied.

Secondly, Neural network doesn’t get the RMSE as well as our expectation. Compared with CNN, neural network itself is not a specific method for image recognition so this would not be surprising. However, we originally expect that neural network may perform better can Regression methods. Now taking a look back, neural network’s failure may be attributed to overfitting and tuning. The method of ’Dropout’ can avoid overfitting but their is no theoretical guidance to the setting of parameters. Beside its high RMSE compared with CNN, it is still time consuming to train neural network. Overall, we do not recommend neural network in face recognition experiment unless perfect parameters can be found.

Thirdly, decision tree is in my recommendation. You may question of my choice because this method doesn’t have access to low RMSE. However, decision tree is very interpretable and visible and it can be used to understand and control the data set. In conclusion, we suggest applying decision tree to the data set first so that you will know what information do the data set conveys. Talking about the adjustments, there is no standard methods here since you can cut or add some branches to the tree as you wish.

Then, we applied a series of linear regression and penalized linear regression. These methods receive ’not bad’ results expect linear regression. In practice, people won’t just use linear regression to a problem unless they can prove a strong linearity. However, we should not abandon penalized linear regression since it is not computational consuming and you can get the result quickly though the code is not run on server. It is hard to make huge improvement because parameters are carefully selected.

Finally, Knn got a surprising result to our expectation. The data set is given in high dimensions, which is the theoretically inappropriate to KNN. The reason might come from the data size. From this experiment, it can be seen that there may exist a relation between dimensions and the amount of data, which need further investigation. However, we still insist not recommending Knn to other high dimension problems like image recognition unless theoretical guidance can be found.

\section{Conclusions}

Overall, we have applied ten different methods to our data set. CNN gets the best RMSE with only 1.972 but it is very time consuming. Decision tree takes advantage of interpretability though it does not have low RMSE. Linear Regression methods are simple enough and only takes a little time to train but they are not recommended for real world applications since we need to pay much more attention to RMSE rather than algorithm complexity.

Neural network and Knn both get surprising result and need further investigation. In this article, we only illustrate some supposes which should be verified in the future.

LBP allows us to capture the details of the image very well, which is important for image processing. Although some results show that LBP may make RMSE worse, most experiments display a good trend. Therefore, LBP is also included as a good solution to this experiment. Moreover, LBP won’t put too much additional information to the data set so that extra computational burden won’t exist.

PCA is a dimension-reduction algorithm, the effect of this method on the data is hard to predict. However, it greatly reduces the dimensions and cut down the consuming time. In this experiment, PCA receives a excellent result. Especially for neural network, this proves our suppose of overfitting, to some degree.

In our experiment, we simply combine LBP and PCA together. However, further improvements can be made by applying these methods independently after a full study of their mechanisms.

\section*{References}
[1] M.Turk\ \& A.Pentland:  Eigenfaces for recognition, {\it J.Cog.Neuroscience}, Volume 3, (1991)

[2] M.Kirby\ \& L.Sirovich:  Application of the Karhunen-loeve procedure for the characterization of human faces, {\it Springer- Verlag, Berlin}, 1989

[3] T.Kohonen:  Self-organization and Associative Memory, {\it IEEE}, 1990

[4] Nitish Srivastava\ \& Geoffrey E. Hinton \ \& Alex Krizhevsky \ \& Ilya Sutskever \ \& Ruslan Salakhutdinov: Dropout: A simple way to prevent neural network from overfitting, {\it Journal of Machine learning Research}, Pages:1929-1958 2015

[5] T.Berg \ \& P.N.Belhumeur:om-vs-pete classifiers and identity-preserving alignment for face verification,{\it 	BMVC}, 2012

[6] Zhu.X \ \& Ramanan D: Face detection, pose estimation, and landmark localization in the wild, {\it CVPR},
 2012
 
 [7] N.Kumar \ \& A.C.Berg \ \& P.N.Belhumeur \ \& S.K.Nayar: Attribute and smile calssifiers for face verification, {\it ICCV}, 2009 
 
 [8] Yaniv Taigman \ \& Ming Yang \ \& Marc’ Aurelio Ranzato: Deep-Face: Closing the Gap to Human-Level Performance, {\it CVPR}, 2014
 
 [9] D.Chen \ \& X.Cao \ \& L.Wang \ \& F.Wen \ \& J.Sun: Bayesian face revisited: A joint formulation, {\it ECCV}, 2012
 
 [10] Y.Taigman \ \& L.Wolf :Leveraging billions of faces to overcome performance barriers in unconstrained face recognition,
 
 [11] Y.Bengio: Learning deep architectures for AI, {\it Foundations and Trends in Machine Learning}, 2009
 
 [12] Y.Sun \ \& X.Wang \ \& X.Tang: Deep convolutional network cascade for facial point detection, {\it CVPR}, 2013
 
 [13] T.Ahonen \ \& A.Hadid \ \& M.pietikainen: Face description with local binary patterns: Application to face recognition, {\it PAMI}, 2006
 
 [14] Karen Simonyan \ \& Andrew Zisserman: Very deep convolutional networks for large-scale image recognition, {\it ICLR}, 2005
 
 [15] M. La Cascia \ \& S. Sclaroff \ \& V. Athitsos: Fast, Reliable Head Track- ing under Varying Illumination: An Approach Based on Registration of Texture-mapped 3D Models.  {\it IEEE Transactions on Pattern Analysis and Machine Intelligence}, April 2000
 
 [16] P. de Cuetos, C. Neti \ \& A. Senior: Audio-visual intent to Speak Detection for Human-computer Interaction {\it In proceedings of the IEEE International Conference on Acoustics, Speech, and Signal Processing} 2000
 
 [17] James M. Rehg \ \& Kevin P. Murphy \ \& Paul W. Fieguth: Vision- based Speaker-detection using Bayesian Networks. {\it In Proceedings of Computer Vision and Pattern Recognition}, Volume 2, pages 110-116, 1999
 
 [18] Y. Matsumoto \ \& A. Zelinsky: An Algorithm for Real-time Stereo Vision Implementation of Head Pose and Gaze Direction Measurement. {\it In IEEE International Conference on Face and Gesture}, page 499, 2000
 
 [19] Second International Workshop on Performance and Evaluation of Tracking and Surveil- lance. {\it IEEE}
  ,December 2001
  
  [20] T. Tan: Second IEEE International Workshop on Visual Surveillance. {\it IEEE}, 1999
  
  [21] Rosalind W. Picard: Affective Computing. {\it MIT Press}, 2009
  
  [22] E. Petajan:  The Communication of Virtual Human Faces using mpeg-4 Tools. {\it In Inter- national Symposium on Circuits and Systems, } Voluume 1, pages 307-310, 2010
  
  [23] Chin-Seng Chua \ \& FengHan \ \& Yeong-KhingHo: 3DHumanFace Recognition using Point Signature. {\it In International Conference on Face and Gesture Recognition}, Voluume 1, pages 307-310, 2010
  
  [24] P. Jonathon Phillips \ \& Patrick J. Rauss \ \& Sandor Z. Der. FERET: (Face Recognition Technology) Recognition Algorithm Development and Test Results. Technical Report ARL-TR-995, {\it Army Research Laboratory}, October, 1996
  
  [25] F. Prokoski: History, Current Status, and Future of Infrared Identification. {\it In Proceedings of IEEE Workshop on Computer Vision Beyond the Visible Spectrum: Methods and Applications}, pages 5-14, June 2000. Facial Thermogram
  
  [26]  P. Jonathon Phillips \ \& Hyeonjoon Moon \ \& Patrick Rauss \ \& Syed A. Rizvi: The FERET September 2009 Database and Evaluation Pro- cedure. {\it In Josef Bigun, Gerard Chollet, and Gunilla Borgefors, editors, Audio- and Video-based Biometric Person Authentication, volume 1206 of Lecture Notes in Computer Science, } pages 395-402. Springer, March 2010.
  
  [27] Duane M.Blackburn \ \& Mike Bone \ \& P.Jonathon Phillips: Facial Recognition Vendor Test 2000 Evaluation Report. {\it Technical Report, Department of Defence Counterdrug Technology Development Program Office, } February 2011
  
  [28] O. Barkan, J. Weill \ \& L. Wolf \ \& H. Aronowitz: Fast high dimen- sional vector multiplication face recognition, {\it ICCV}, 2013
  
  [29] X. Cao \ \& D. Wipf \ \& F. Wen \ \& G. Duan \ \& J. Sun: A practical transfer learning algorithm for face verification, {\it ICCV}, 2013
  
  [30] D.Chen \ \& X.Cao \ \& F.Wen \ \& J.Sun: Blessing of dimensionality: High- dimensional feature and its efficient compression for face verification, {\it CVPR}, 2013
  
  [31] M.Osadchy \ \& Y.LeCun \ \& M.Miller: Synergistic face detection and pose estimation with energy-based models, {\it JMLR}, 2007
  
  [32] S. Chopra \ \& R. Hadsell \ \& Y. LeCun: Learning a similarity metric discriminatively, with application to face verification, {\it CVPR}, 2005
  
  [33] G. B. Huang \ \& H. Lee \ \& E. Learned-Miller: Learning hierarchical representations for face verification with convolutional deep belief networks, {\it CVPR}, 2012
  
  [34] Jochen Triesch \ \& Christoph von der Malsburg: Fisher vector faces in the wild, {\it BMVC}, 2013
  
  [35] K. Simonyan \ \& O. M. Parkhi \ \& A. Vedaldi \ \&  A. Zisserman: elf-organized Integration of Adaptive Visual Cues for Face Tracking. {\it In International Conference on Face and Gesture Recognition, IEEE}, 2015
  
  [36]  G. J. Edwards \ \& C. J. Taylor \ \& T. F. Cootes: Interpreting Faces using Active Appearance Models. {\it In International Conference on Face and Gesture Recognition}, April 2012
  
  [37] Roberto Brunelli \ \& Tomaso Poggio: Face Recognition: Features versus Templates. {\it IEEE Transactions on Pattern Analysis and Machine Intelligence}, October 2013
  
  [38] Ian Craw \ \& Peter Cameron: Face Recognition by Computer {\it Proceedings of the British Machine Vision Conference, Springer Verlag}, 2012
  
  [39] G.J.Edwards \ \& C.J.Taylor \ \& T.F.Cootes: Interpreting Faces using Active Appearance Models. {\it In International Conference on Face and Gesture Recognition },April 2012
  
  [40] Christopher M.Bishop: Pattern recognition and Machine learning, {\it Springer} 2006
\end{document}